\documentclass[11pt, a4paper, logo, twocolumn, address]{deepmind}

\usepackage{tocbibind}
\usepackage[all]{hypcap}
\usepackage[authoryear, sort&compress, round]{natbib}
\usepackage{relsize}
\newcommand{\Cpp}{\texorpdfstring{C\kern-0.05em\protect\raisebox{.35ex}{\textsmaller[2]{+\kern-0.05em+}}}{C++}}

\newcommand*{\figref}[1]{\hyperref[{fig:#1}]{Fig.~\ref*{fig:#1}}}

\title{DeepMind Lab2D}

\reportnumber{} 

\renewcommand{\today}{November 2020} 

\author[1]{Charles Beattie}
\author[1]{Thomas K\"{o}ppe}
\author[1]{Edgar A. Du\'e\~nez-Guzm\'an}
\author[1]{Joel Z. Leibo}

\affil[1]{DeepMind}

\correspondingauthor{Charles Beattie $\langle$cbeattie@google.com$\rangle$}

\begin{abstract}
We present \emph{DeepMind Lab2D}, a scalable environment simulator for artificial intelligence research that facilitates researcher-led experimentation with environment design. DeepMind Lab2D was built with the specific needs of multi-agent deep reinforcement learning researchers in mind, but it may also be useful beyond that particular subfield.
\end{abstract}

\begin{document}

\maketitle

\section{Introduction}

Are you a product of your genes, brain, or environment? What will an artificial intelligence (AI) be? The development of AI systems is inextricably intertwined with questions about the fundamental causal factors shaping any intelligence, including natural intelligence. Even in a completely abstract learning scenario, prior knowledge is needed for effective learning, but prior learning is the only realistic way to generate such knowledge. 

The centrality of this dynamic is illustrated by experiments in biology. In laboratory animals, manipulations of the rearing environment produce profound effects on both brain structure and behavior. For instance, laboratory rodent environments may be enriched by using larger cages which contain larger groups of other individuals---creating more opportunities for social interaction, variable toys and feeding locations, and a wheel to allow for the possibility of voluntary exercise. Rearing animals in such enriched environments improves their learning and memory, increases synaptic arborization, and increases total brain weight~\citep{van2000neural}. It is the interaction of environmental factors that is thought to produce the enrichment effects, not any single factor in isolation.

While it is conceivable that AI research could stumble upon a perfectly general learning algorithm without needing to consider its environment, it is overwhelmingly clear that environments affect learning in ways that are not just arbitrary quirks, but real phenomena which we must strive to understand theoretically. There is structure there. For instance, the question of why some environments only generate trivial complexity, like tic-tac-toe, while others generate rich complexity, like Go, is not just a matter of the state space size. It is a real scientific question that merits serious study. We have elsewhere referred to this as \emph{the problem problem}~\citep{leibo2019autocurricula}.

One reason it is currently difficult to undertake work aimed at the problem problem is that it is rare for any individual person to possess expertise in all the relevant areas. For instance, machine learning researchers know how to design well-controlled experiments but struggle with the necessary skills that are more akin to computer game design and engineering. Another reason why it is difficult to pursue this hypothesis is the prevailing culture in machine learning that views any tinkering with the environment as ``hand-crafting'', ``special-casing'', or just ``cheating''. These attitudes are misguided. In our quest for generality, we must not forget the great diversity and particularity of the problem space.

A diverse set of customizable simulation environments for large-scale 3D environments with varying degrees of physical realism exist~\citep{todorov2012mujoco, kempka2016vizdoom, beattie2016deepmind, leibo2018psychlab, juliani2018unity}. For 2D, excellent simulation environments also exist \citep{stepleton2017pycolab,zheng2019magent,gym_minigrid,LanctotEtAl2019OpenSpiel,shuo2019maenvs,suarez2019neural,platanios2020jelly,schaul2013video}; however, they fell short, at the time this project began, in at least one of our requirements of composability, flexibility, multi-agent capabilities, or performance.

\subsection{DeepMind Lab2D}

\emph{DeepMind Lab2D} (or ``DMLab2D'' for short)\footnote{\url{https://github.com/deepmind/lab2d}} is a platform for the creation two-dimensional, layered, discrete ``grid-world'' environments, in which \emph{pieces} (akin to chess pieces on a chess board) move around. This system is particularly tailored for multi-agent reinforcement learning. The computationally intensive engine is written in \Cpp{} for efficiency, while most of the level-specific logic is scripted in Lua.

\paragraph{The grid.}
The environments of DMLab2D consist of one or more layers of two-dimensional grids. A position in the environment is uniquely identified by a coordinate tuple $(x, y, \text{layer})$. Layers are labeled by strings, and the $x$- and $y$-coordinates are non-negative integers. An environment can have an arbitrary number of layers, and their rendering order is controlled by the user.

\paragraph{Pieces.}
The environments of DMLab2D are populated with pieces. Each piece occupies a position $(x, y, \text{layer})$, and each position is occupied by at most one piece. Pieces also have an orientation, which is one of the traditional cardinal directions (north, east, south, west). Pieces can move around the $(x, y)$-space and reorient themselves as part of the evolution of the environment, both relatively to their current position/orientation and absolutely. It is also possible for a piece to have no position, in which case it is ``off the board''. Pieces cannot freely move among layers; instead, a piece's layer is controlled through its state (described next).

\paragraph{States.}
Each piece has an associated \emph{state}. The state consists of a number of key-value attributes. Values are strings or lists of strings. The possible values are fixed by the designer as part of the environment. The state of each piece can change as part of the evolution of the environment, but the state change can only select from among the fixed available values.

The state of a piece controls the piece's appearance, layer, group membership, and behavior. Concretely, the state of a piece comprises the following attributes:
\begin{itemize}
\item \texttt{layer} (string):
    The label of the layer which the piece occupies.
\item \texttt{sprite} (string):
    The name of the sprite used to draw this piece.
\item \texttt{groups} (list of strings):
    The groups of which this piece is a member. Groups are mostly used
    for managing updater functions.
\item \texttt{contact} (string):
    A tag name for a contact event. Whenever the piece enters (or leaves)
    the same $(x, y)$-coordinate as another piece (which is necessarily on
    a different layer), all involved pieces experience a contact event.
    The event is tagged with the value of this attribute.
\end{itemize}
An attempt to change a piece's state fails if the piece's resulting $(x, y, \text{layer})$-position is already occupied.

\paragraph{Callbacks.}
Most of the logic in an environment is implemented via \emph{callbacks} for specific types (states) of pieces. Callbacks are functions which the engine calls when the appropriate event or interaction occurs.

\paragraph{Raycasts and queries.}
The engine provides two ways to enumerate the pieces in particular positions (and layers) on the grid: \emph{raycasts} and \emph{queries}. A raycast, as the name implies, finds the first piece, if any, in a ray from a given position. A query finds all pieces within a particular area in the grid, shaped like a disc, a diamond, or a rectangle.

\subsection{Why 2D, not 3D?}

Two-dimensional environments are inherently easier to understand than three-dimensional ones, at very little, if any, loss of expressiveness. Even a game as simple as Pong, which essentially consists of three moving rectangles on a black background, can capture something fundamental about the real game of table tennis. This abstraction makes it easier to capture the essence of the problems and concepts that we aim to solve. 2D games have a long history of making challenging and interpretable benchmarks for artificially intelligent agents~\citep{shannon1950chess, samuel1959some, mnih2015human}.

Rich complexity along numerous dimensions can be studied in 2D just as readily as in 3D, if not more so. When studying a particular research question, it is not clear \emph{a priori} whether specific aspects of 3D environments are crucial for obtaining the desired behavior in the training agents. Even when explicitly studying phenomena like navigation and exploration, where organisms depend on complex visual processing and continuous-time physical environments, researchers in reinforcement learning often need to discretize the interactions and observations so that they become tractable. Moreover, 2D worlds can often capture the relevant complexity of the problem at hand without the need for continuous-time physical environments. This pattern where studying phenomena on 2D worlds has been a critical first step towards further advances in more complex and realistic environments is ubiquitous in the field of artificial intelligence. 2D worlds have been successfully used to study problems as diverse as social complexity, navigation, imperfect information, abstract reasoning, exploration, and many more~\citep{leibo2017multi, lerer2017maintaining, zheng2020ai, rafols2005using, ullman2009help}.

Another advantage of 2D worlds is that they are easier to design and program than their 3D counterparts. This is particularly true when the 3D world actually exploits the space or physical dynamics beyond the capabilities of 2D ones. 2D worlds do not require complex 3D assets to be evocative, nor do they require reasoning about shaders, lighting, and projections. In most 2D worlds, the agent's egocentric view of the world is inherently compatible with the allocentric view (i.e. the third-person or world view). That is, typically the agent's view is simply a movable window on the whole environment's view.

In addition, 2D worlds are significantly less resource-intensive to run, and typically do not require any specialized hardware (like GPUs) to attain reasonable performance. This keeps specialized hardware, if any, available exclusively for the intensive work of training the agents. Using 2D environments also enables better scalability to a larger number of agents interacting with the same environment, as it costs only very little to render another agent's view. Running 2D simulations is within the capabilities of smaller labs, whereas most 3D physics-based reinforcement learning is still prohibitively expensive in many settings.

\subsection{Multi-player support and benchmarking with human players}

A large fraction of human skills are social skills. To probe these, simulation environments must provide robust support for multi-agent systems. Most existing environments, however, only provide poor support for multiple players.

DeepMind Lab2D supports multiple simultaneous players interacting in the same environment. These players may be either human or computer-controlled, and it is possible to mix human and computer-controlled players in the same game.

Each player can have a custom view of the world that reveals or obscures particular information, controlled by the designer. A global view, potentially hidden from the players, can be set up and can include privileged information. This can be used for imperfect information games, as well as for human behavioral experiments where the experimenter can see the global state of the environment as the episode is progressing.

\subsection{Exposing metrics, supporting analysis}

DeepMind Lab2D provides several flexible mechanisms for exposing internal environment information: The simplest form is through \emph{observations}, which allow the researcher to add specific information from the environment to the observations that are produced at each time step. The second way is through \emph{events}, which, similar to observations, can be raised from within the Lua script. Unlike observations, events are not tied to time steps but instead are triggered on specific conditions. Finally, the \emph{properties} API provides a way to read and write parameters of the environment, typically parameters that change rarely.

\section[Example results]{Example results in\\deep reinforcement learning}

\begin{figure}[tb]
    \centering
    \includegraphics[width=\columnwidth]{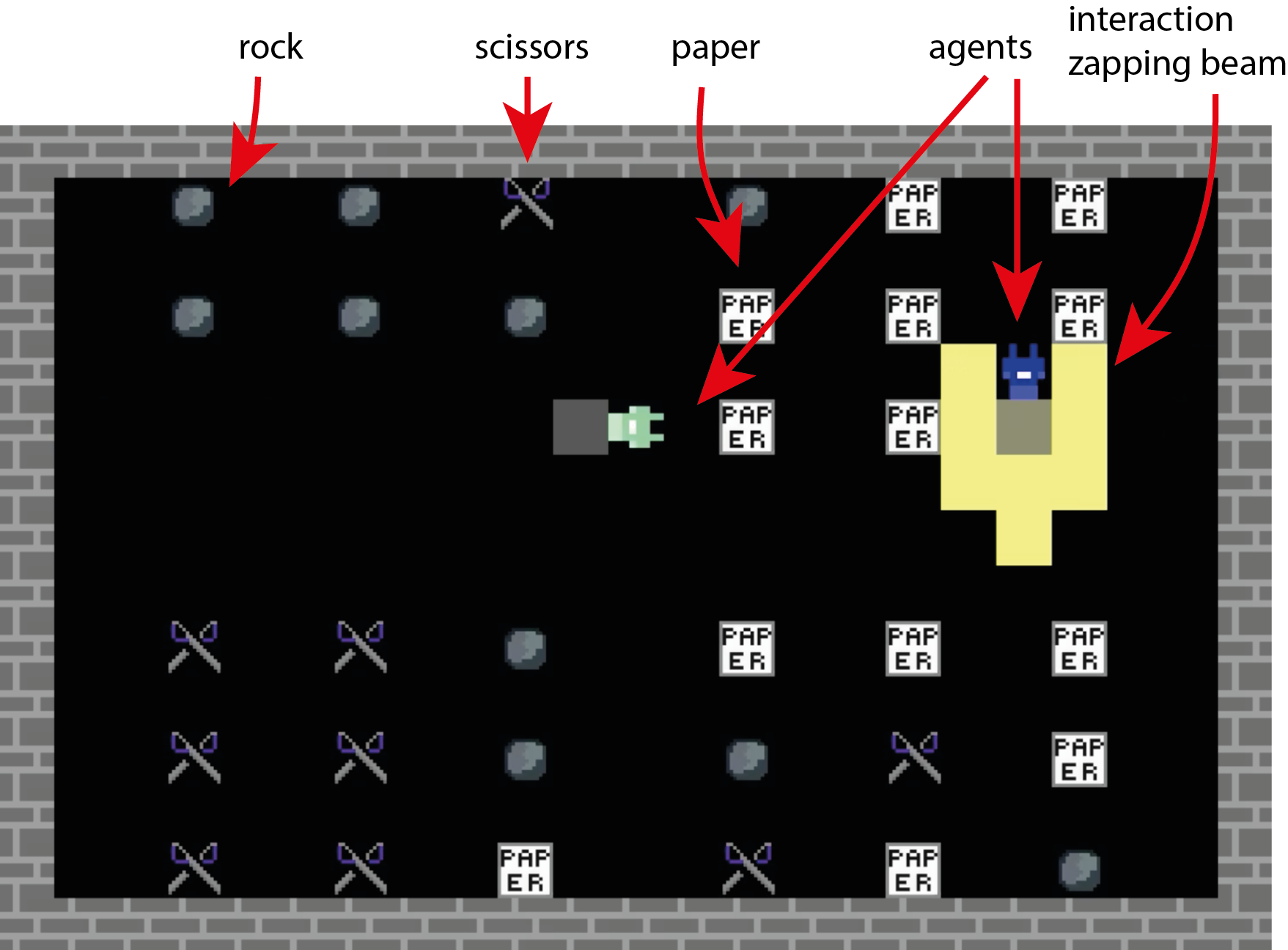}
    \caption{``Running With Scissors'' screenshot.}\label{fig:screenshot}
\end{figure}

\begin{figure*}[t]
    \centering
    \includegraphics[width=\textwidth]{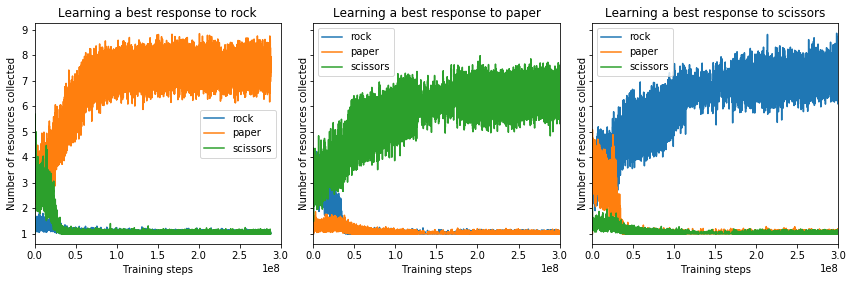}
    \caption{A new agent implementing the advantage actor-critic algorithm  can be trained to best respond to frozen agents implementing ``semi-pure'' strategies~\citep{mnih2016asynchronous}.}\label{fig:best_responses}
\end{figure*}

For an example, we consider a game called ``Running With Scissors'' (\figref{screenshot}). A variant of this game with simpler graphics was first described in \cite{vezhnevets2020options}. It can be seen as a spatially and temporally embedded extension to Rock-Paper-Scissors. As such, it inherits the rich game-theoretic structure of its parent (described in e.g.~\cite{weibull1997evolutionary}). The main difference is that, unlike in the matrix game, agents in Running With Scissors do not select their strategies as atomic decisions. Instead, they must learn \emph{policies} to implement their strategic choices. They must decide \emph{how} to ``play rock'' (or paper, or scissors), in addition to deciding \emph{that} they should do so. Furthermore, it is possible---though not trivial---to observe the policy one's partner is starting to implement and take countermeasures. This induces a wealth of possible feinting strategies, none of which could easily be captured in the classical matrix game formulation.

Agents can move around the map and collect resources: rock, paper, and scissors. The environment is $16 \times 24$ units in size but agents view it through a $5 \times 5$ window.\footnote{This partial viewing window is a square. The agent sees 3 rows in front of itself, 1 row behind, and 2 columns to either side. Elementary actions are to move forward, backward, strafe left, strafe right, turn left or turn right, and fire the interaction beam. A gameplay video may be viewed at \url{https://youtu.be/IukN22qusl8}.} The episode ends either when a timer runs out or when there is an interaction event, triggered by one agent zapping the other with a beam. The resolution of the interactions is driven by a traditional matrix game, where there is a payoff matrix describing the reward produced by the pure strategies available to the two players. In Running With Scissors, the zapping agent becomes the row player and the zapped agent becomes the column player. The actual strategy of the player depends on the resources it has picked up before the interaction. These resources are represented by the resource vector
\[ v \in \Delta^2 \subset \mathbb{R}_{\text{rock}} \oplus \mathbb{R}_{\text{paper}}
         \oplus \mathbb{R}_{\text{scissors}} \text{, } \]
where
\[ \Delta^2 \coloneq \bigl\{ (x_1, x_2, x_3) : 0 \leq x_1, x_2, x_3
   \text{ , } \textstyle\sum_{i} x_i = 1 \bigr\}  \]
is the standard 2-simplex. The initial value of the vector is the centroid $\bigl(\frac{1}{3}, \frac{1}{3}, \frac{1}{3}\bigr)^T$. The more resources of a given type an agent picks up, the more committed the agent becomes to the pure strategy corresponding to that resource. The rewards $r_\text{row}$ and $r_\text{col}$ for the (zapping) row and the (zapped) column player, respectively, are assigned via
\begin{equation*}
    r_{\text{row}} = v^T A \, v
                   = -r_{\text{col}} \text{ ,}
\end{equation*}
where
\[ A =
\begin{bmatrix}
\hphantom{+}0 & -1 & +1\\
+1 & \hphantom{+}0 & -1\\
-1 & +1 & \hphantom{+}0
\end{bmatrix}
\text{ .} \]

To obtain high rewards, an agent should correctly identify what resource its opponent is collecting (e.g.\ rock) and collect the resource corresponding to its counter-strategy (e.g.\ paper). In addition, the rules of the game, i.e.\ the dynamics of the environment are not assumed known to the agents. They must explore to discover them. Thus Running With Scissors is simultaneously a game of imperfect information---each player possesses some private information not known to their adversary (as in e.g.\  poker~\citep{sandholm2015solving})---and incomplete information, lacking common knowledge of the rules~\citep{harsanyi1967games}.

In Rock-Paper-Scissors, when faced with an opponent playing rock, a best responding agent will come to play paper. When faced with an opponent playing paper, a best responding agent will learn to play scissors. And, when one's opponent plays scissors, one will learn to counter with rock. \figref{best_responses} shows that the same incentives prevail in Running With Scissors. However, in this case, the policies to implement strategic decisions are more complex since agents must learn to run around the map and collect the resources. Much more complex policies involving scouting and feinting can also be learned in this environment. See~\cite{vezhnevets2020options} for details.

To give an indication of the simulation performance, two random agents playing against one another, receiving full RGB observations of size $80 \times 80$px ($16 \times 16$px per tile, with $5 \times 5$ tiles), run with an average of 250,000 frames per second (measured over 1000 episodes of 1000 steps each), on a single core of an Intel Xeon W-2135 (``Skylake'') CPU at 3.70GHz. The training example shown in \figref{best_responses} took several days to complete, and the cost of running the simulation is thus entirely negligible.

\section{Discussion}

Artificial intelligence research based on reinforcement learning is beginning to mature as a field. The need for rigorous standards by which the correctness, scale, reproducibility, ethicality, and impact of a contribution may be assessed is now accepted~\citep{osband2019behaviour, mitchell2019model, henderson2020towards, khetarpal2018re}. 

But in all these well-received calls for rigor in AI, the humble simulation environment gets shortchanged. It would appear that many researchers consider the environment to be none of their concern. A more holistic (and realistic!) view of their work suggests otherwise. Research workflows involve significant time spent authoring game environments and intelligence tests, adding analytic methods, and so forth. But these activities are usually not as simple and easy to extend as they ought to be, though they are clearly critical to the success of the enterprise.

We think that progress toward artificial general intelligence requires robust simulation platforms to enable \emph{in silico} exploration of agent learning, skill acquisition, and careful measurement. We hope that the system we introduce here, DeepMind Lab2D, can fill this role. It generalizes and extends a popular internal system at DeepMind which supported a large range of research projects. It was especially popular for multi-agent research involving workflows with significant environment-side iteration. In our own experience, we have found that DeepMind Lab2D facilitates researcher creativity in the design of learning environments and intelligence tests. We are excited to see what the research community uses it to build in the future.

\section{Acknowledgements}

We thank the following people for their contributions to this project: 
Antonio Garc{\'\i}a Casta{\~n}eda,
Edward Hughes, 
Ramana Kumar, 
Jay Lemmon,
Kevin \mbox{McKee},
Haroon Qureshi,
Denis Teplyashin,
V\'ictor Vald\'es, and
Tom Ward.


\balance

\bibliography{main}

\end{document}